\documentclass[11pt]{article}

\usepackage{natbib}

\usepackage[pdftex]{graphicx}
\usepackage[pdftex]{color}

\newcommand{\figref}[1]{Figure~\ref{#1}}
\newcommand{\tabref}[1]{Table~\ref{#1}}

\catcode`\@=11
\def\@listI{\leftmargin=\leftmargini
\partopsep=0pt \parsep=0pt \topsep=0pt \itemsep=0pt \relax}
\def\@listii{\leftmargin=\leftmarginii \labelwidth=\leftmarginii
\advance\labelwidth by-\labelsep
\parsep=0pt \topsep=0pt \itemsegp=0pt \relax}
\let\@listi\@listI \@listi

\catcode`\@=12

{\begin{itemize}\setlength{\parsep}{0pt}\setlength{\itemsep}{0pt}%
\setlength{\topsep}{0pt}\setlength{\parskip}{0pt}\setlength{\partopsep}{0pt}}%
{\end{itemize}}

{\begin{enumerate}\setlength{\parsep}{0pt}\setlength{\itemsep}{0pt}%
\setlength{\topsep}{0pt}\setlength{\parskip}{0pt}\setlength{\partopsep}{0pt}}%
{\end{enumerate}}

{\begin{description}\setlength{\parsep}{0pt}\setlength{\itemsep}{0pt}%
\setlength{\topsep}{0pt}\setlength{\parskip}{0pt}}%
{\end{description}}

\title{Word Familiarity and Frequency}

\author{Kumiko Tanaka-Ishii \hspace*{1cm} Hiroshi Terada \\
Graduate School of Information Science and Technology \\
University of Tokyo \\
7-3-11 Hongo Bunkyoku, Japan.    \\
{\tt \{kumiko, hiroshiTerada\}@i.u-tokyo.ac.jp}\\
tel\&fax +81-3-5209-3530 
}

\date{}

\begin{document}

\maketitle

\begin{abstract} 
Word frequency is assumed to correlate with word familiarity, but the strength of this 
correlation has not been thoroughly investigated. In this paper, we report on our analysis 
of the correlation between a word familiarity rating list obtained through a 
psycholinguistic experiment and the log-frequency obtained from various corpora of 
different kinds and sizes (up to the terabyte scale) for English and Japanese. Major 
findings are threefold: First, for a given corpus, familiarity is necessary for a word to 
achieve high frequency, but familiar words are not necessarily frequent. Second, 
correlation increases with the corpus data size. Third, a corpus of spoken language 
correlates better than one of written language. These findings suggest that cognitive 
familiarity ratings are correlated to frequency, but more highly to that of spoken rather 
than written language. 
\end{abstract}

\section{Introduction} 
\label{sec:introduction}
Word familiarity is the relative ease of perception attributed to every word. For example, 
the two words ``encounter'' and ``meeting'' could be used in a similar way, but 
``meeting'' is cognitively easier than ``encounter''. Word familiarity is interesting from a 
scientific viewpoint to investigate the mental process of word-meaning acquisition. It is 
also related to language engineering where it is applied in language education and 
e-learning.

Within the past few decades, attempts have been made to measure word familiarity 
through human experiments within the psycholinguistic domain. Studies have generated 
several word familiarity lists such as Wilson's list \citep{wilson88} and the MRC 
database \citep{mrc} in English, which consists of several thousand words with 
familiarity ratings. In Japanese, Amano's list \citep{ntt00} contains about 70,000 pairs 
of words and corresponding ratings.

Even though such lists have been generated, it is still not entirely clear what processes 
are involved when readers rate the familiarity of a word. Familiarity ratings have often 
been interpreted as a measure of the frequency of exposure to a word \citep{mrc}. 
Many studies show that a word's frequency affects its perception \citep{segui82} 
\citep{dupoux90} \citep{marslen90}, while some studies point out that for word 
perception the familiarity thus acquired through experiment is a better predictor than 
frequency \citep{gernsbacher84} \citep{gordon85} \citep{kreuz87} 
\citep{nusbaum84}. A psychological study reports the relation between
word familiarity and frequency effect in visual and auditory word recognition
through experiments \citep{connine90}.

All such previous work was done on the psycholinguistic side and there have been only 
limited attempts on the computational linguistic side to verify the relationship between 
language statistics and familiarity. This paper reports our investigation of the relation 
between the word familiarity ratings obtained in a psycholinguistic experiment and the 
word frequency acquired from various corpora. Our study is limited to investigating the 
relation through statistical measurements found in corpora, and does not include any 
psycholinguistic or cognitive experimental results. After showing the basic degree of 
correlation and its general characteristics, we discuss how the corpus data size and 
corpus type affect the degree of correlation.

Our study is intended to contribute to a better scientific understanding of word 
familiarity from a linguistic data viewpoint. The correlation of familiarity with the log 
frequency of use --- information carried by the word--- is measured per, in a global 
sense, Weber-Fechner's law, which states that the relationship between stimulus and 
perception is generally logarithmic \citep{weber06} \citep{fechner04}. Familiarity can 
therefore be a linguistic trace under this law; or, to put it otherwise, Weber-Fechner's 
law can be considered a general model underlying familiarity. This article attempts to 
consider the contours of the nature of familiarity through an empirical approach.

The findings in this article should contribute to language engineering, especially to 
language education, since an accurate list of word familiarity ratings is the key to 
reliable reading level assessment. So far, every method we know of measures 
vocabulary difficulty using either a frequency count or a vocabulary list of easy words. 
For example, the Dale-Chall method \citep{chall95}, counts the number of words in a 
text which are not registered in the Dale-Chall list of 3000 basic words. This list 
indicates that ``easy'' words appear in most texts, but the lists are manually constructed 
without a solid scientific grounding. Familiarity ratings have a great potential to replace 
such a list, but the experimental workload needed to generate such lists in various 
languages is a problem. Thus, a statistical measure that can be used as a {\em 
pseudo}-familiarity rating would be valuable. As for the use of frequency in reading 
level assessment, recent studies have attempted to measure vocabulary difficulty by 
means of frequency \citep{kevyn04} \citep{schwarm05}. However, the function of 
frequency with respect to reading level has not been clarified in these studies. Thus, 
investigating the nature of frequency with respect to familiarity ratings could provide 
the key to a better reading level assessment. Our intention in this paper is to show that 
the log-frequency of a word's use is a possible measure of such pseudo-familiarity, if it 
is measured in an adequate corpus. We investigate which corpus conditions lead to 
higher correlation between a word's log-frequency and its familiarity.

\begin{table}[t] 
\begin{center} 
\caption{Our Database Used to Measure Frequency} 
\label{tab:corpus} 
\begin{tabular}{|l|r|r|c|p{3cm}|} 
\hline label & Num. tokens & Num. types & spoken/ & kind of \\
& (total words) & (different words) & written & text \\ 
\hline 
\hline 
WSJ (3-years) & 42287431 &127353 & written & newspaper \\
Wikipedia-E & 711143194 &168533 & written & encyclopedia \\
Web-E & 88267343947 &204724587 & written & mixture \\
BNC & 97098970 & 364262 & both & mixture \\
MICASE & 1279792 & --- & spoken & academic speaking \\
\hline 
\hline 
Mainichi (5 years)& 80709011 &198767 & written & newspaper \\
Wikipedia-J & 130418600 &619636 & written & encyclopedia \\
Web-J & 7183558565 &5474644 & written & mixture \\
Aozora & 25975560 &139961 & written & literature \\
Spoken Corpus-J (SCJ) & 7498763 &47767 & spoken & academic lecture \\
\hline 
\end{tabular} 
\end{center} 
\end{table}

\section{Database} 
\subsection{Word Familiarity Lists}
Most readers would probably agree that ``meeting'' is more familiar than ``encounter''. 
Likewise, coherency is assumed among adults regarding the relative familiarity of 
words. The exact definition of {\em familiarity} remains controversial and the term is 
not well-defined. Moreover, no psycholinguistic experiment is free from individual 
variation. Still, familiarity rating attempts to extract such cohesion from human thought 
through psycholinguistic experiments by taking the view that familiarity lists are a 
database which partly reveals human perception about language. In general, familiarity 
ratings are obtained by asking people to subjectively score the familiarity of words on 
multiple levels, and then post-processing the scores in some standard way defined in 
psychological experiment methodologies \citep{colheart81a} \citep{wilson88} 
\citep{amano95}. The signals were given word based without context.

In English, a list of several thousand words was first reported in \citep{nusbaum84}. A 
more recent effort has been the MRC project reported in \citep{cortheart81}. In 
addition to familiarity, the list contains various scores for each word, such as the 
acquisition age and the meaningfulness. In total, the MRC list contains 150,837 words 
for 26 different linguistic properties. However, each score is given for only part of this 
word set and the familiarity rating is available for only 4894 words through the MRC 
web-based interface \citep{mrc}. This list is used in our study and is called the {\em 
MRC} list in this article. The MRC list was constructed by merging three different sets 
of familiarity norms: Pavio (unpublished), \cite{toglia78} and \cite{gilhooly80}. Each 
database was constructed separately by asking subjects to rate words by familiarity 
levels. The three sets of norms were then merged under statistical consideration as 
described in more detail in Appendix 2 of the MRC Psycholinguistic Database User 
Manual \citep{colheart81a}. The historical consequence of this mixture possibly makes 
the use of this database for this research questionable, but this is the only large 
familiarity score available in English to date.

For Japanese, \cite{amano08} generated a word familiarity list of 68,550 words 
through a large-scale experiment \citep{ntt00}. This list, referred to as the {\em 
Amano} list, can be purchased and was used in our study. The values were obtained by 
asking 40 people to score content words in a standard Japanese dictionary, with a 
familiarity score of 7 different levels. Each person scored 9000 words, and only 
statistically plausible judgments were used to generate the final score of familiarity 
\citep{amano08}.

In both of these lists, the familiarity rating ranges from 1.0 to 7.0, with 7.0 being the 
most familiar and 1.0 being the least familiar\footnote{ In MRC, the original ratings 
ranged from 100 to 700; we divided these by 100 in our experiment so that they would 
be consistent with the Amano list.}. The two lists differ in that the Amano list only 
contains {\em content} words, whereas MRC also has some functional words. Note that 
functional words can be familiar or unfamiliar. For example, the familiarity of ``must'' is 
clearly higher than that of ``ought''. The functional words in the MRC list do not 
necessarily contain most frequently used functional words such as ``and'' and ``to''.

The most familiar and unfamiliar words taken from the MRC and Amano lists are listed 
in the first column of \tabref{tab:ex1} for English and \tabref{tab:ex2} for Japanese. 
For Japanese, the word meaning is shown under the word only for content words. Words 
such as daily and greeting words are among the most familiar words, whereas words 
which are difficult to find even in dictionaries are among the most unfamiliar. For 
Japanese, especially, even though the words might not seem difficult from the English 
translations, the unfamiliarity of words in the lower block of \tabref{tab:ex2} is 
prominent, which is partly noticeable from the excessive complexity of the characters.

\begin{table}[t] 
  \begin{center} 
    \caption{Familiar and frequent words (top), unfamiliar and rare words (bottom) in English} 
 \label{tab:ex1} 
 \begin{tabular}{|l||l||l|l|l|l|l|} 
 \hline & \multicolumn{1}{c||}{MRC} & \multicolumn{1}{|c|}{WSJ} & \multicolumn{1}{|c|}{BNC} & 
 \multicolumn{1}{|c|}{Wikipedia-E} & \multicolumn{1}{|c|}{Web-E}& 
 \multicolumn{1}{|c|}{MICASE}\\
 \hline 
 Most &breakfast&the&the&on&the&the \\
 2nd &afternoon&be&be&as&and&that \\
 3rd &clothes&of&of&for&be&and \\
 4th &dad&to&and&to&of&you \\
 5th &bedroom&a&to&a&to&I \\
 6th &girl&in&a&in&a&of \\
 7th &radio&and&in&and&in&it \\
 8th &book&say&have&be&for&to \\
 9th &water&that&that&of&you&a \\
 10th &newspaper&have&it&the&I&in \\
 \hline 
 \hline & \multicolumn{1}{c||}{MRC} & \multicolumn{1}{|c|}{WSJ} & \multicolumn{1}{|c|}{BNC} & 
 \multicolumn{1}{|c|}{Wikipedia-E} & \multicolumn{1}{|c|}{Web-E}& 
 \multicolumn{1}{|c|}{MICASE}\\
 \hline 
 Most &metis&zonal&paulet&fatly&sandarach&abatement \\
 2nd &anele&ziti&shivery&fading&monopetalous&amuse \\
 3rd &goral&zestful&decoction&exorable&palatially&angel \\
 4th &pavis&yodel&folate&edgily&febricity&clown \\
 5th &witan&yeti&sural&bronchitic&coriss&hobby \\
 6th &jupon&dater&corse&brassart&aeilderts&melon \\
 7th &kevel&twinbed&armor&benthonic&sulphurize&oasis \\
 8th &lagan&daby&broil&blushful&amortizement&noodle \\
 9th &daman&voiron&trey&bawdily&ruralize&prefix \\
 10th &manus&toxemia&velveteen&alsoran&allotropism&token \\
 \hline
\end{tabular} 

  \end{center} 
\end{table}

\begin{table}[t] 
\footnotesize
  \begin{center} 
    \caption{Familiar and frequent words (top), unfamiliar and rare words (bottom) in Japanese}
 \label{tab:ex2} 
 \includegraphics[width=\textwidth]{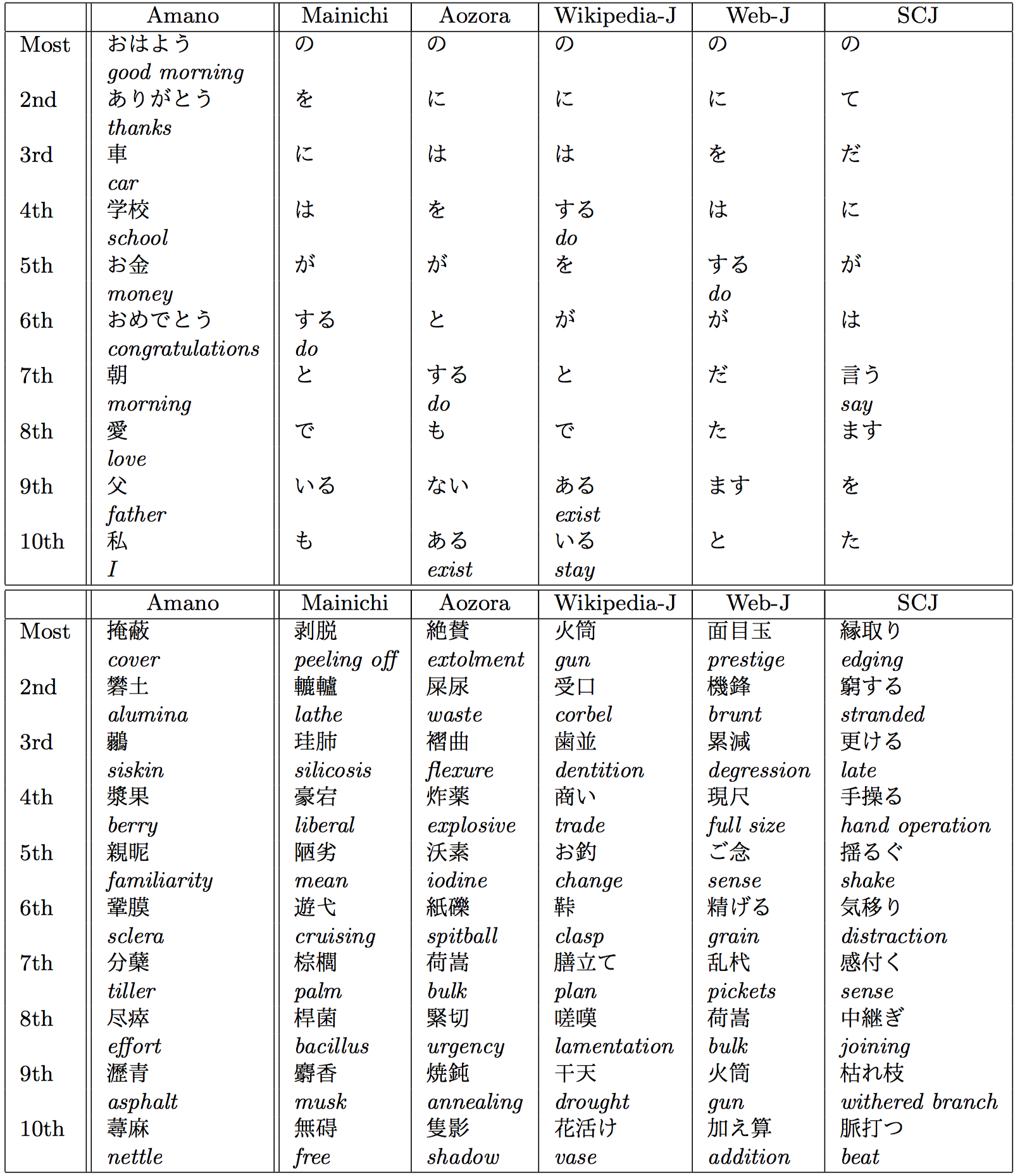}
\vspace*{2cm}
\end{center}% \vspace*{2cm} 
\end{table}

\subsection{Corpora}
Data listed in \tabref{tab:corpus} are used to measure frequency. The first block shows 
corpora for English and the second block shows those for Japanese. Our data includes 
newspaper corpora, data obtained from Wikipedia, web data, mixed-type corpora, and 
spoken language corpora.

As the newspaper corpora, we used the three-year WSJ newspaper corpus for English 
and the five-year Mainichi newspaper corpus for Japanese.

From Wikipedia, we extracted all English and Japanese texts and eliminated tags to 
acquire plain data. We obtained 1,912,595 pages for English, which amounted to 4.74 
gigabytes without tags, whereas we obtained 372,890 pages for Japanese, which 
amounted to 945 megabytes.

The Web data was crawled\footnote{We thank Associate Professor K. Taura, of the 
University of Tokyo for offering us this data.} and downloaded from the Internet in 
autumn 2006. Here, too, markup tags were eliminated and texts in English and Japanese 
were extracted. For English texts, 265,823,502 pages were scanned, which amounted to 
1.9 terabytes of text data without tags, and for Japanese 12,751,271 pages were scanned 
and 69 gigabytes of data obtained.

BNC was used as a mixed-type corpus for English, while Aozora, a collection of 
literature corresponding to the Gutenberg project in Japanese, was used for Japanese. 
Spoken language corpora transcribed from spoken recordings were also used in our 
studies. For English we used MICASE (Michigan Corpus of Academic Spoken 
English)\footnote{ {\tt http://quod.lib.umich.edu/m/micase/}. Since MICASE was only 
available via a web-based interface, the statistics shown in \tabref{tab:corpus} were 
taken from the MICASE web page.}, which is available only online. For Japanese, we 
used the Spoken Corpus of Japanese.

For every corpus except MICASE, frequency was measured after lemmatizing each 
word into standard forms using 
Tree-tagger\footnote{ http://www.ims.uni-stuttgart.de/projekte/corplex/TreeTagger/} 
for English and Chasen\footnote{ http://chasen-legacy.sourceforge.jp/ } for Japanese. 
The standard form was used since every word in the familiarity list is in the standard 
form. Words of the same form with different senses are not distinguished, since as noted 
at the beginning of the previous section, a familiarity score was also measured without 
such disambiguation by context. Note that even after processing words into their 
standard forms, frequency lists acquired from large corpora can be huge. For example, 
the list for Web-E was more than 2 gigabytes.

From all corpora, the most frequent words and randomly chosen words which acquired 
the lowest frequency of $n$ ($n$ = 1\ldots 10) are listed in \tabref{tab:ex1} and 
\tabref{tab:ex2}. In both tables, the most frequent words are common to most corpora 
in each language, but they do not match the most familiar words. Here, we see an 
obvious difference between frequency and familiarity: most familiar words are content 
words, whereas the most frequent words are functional, which is probably due to the 
lack of functional words in the familiarity list. The same tendency also holds in 
Japanese. As for the most rare and unfamiliar words, none of the lists share words, 
which is to be expected since language is characterized by a {\em large number of rare 
events} (LNRE).

\section{The basic correlation}
When a corpus is large, the frequency of a word will obviously be high. In contrast, the 
word familiarity rating ranges from 1.0 to 7.0. If a correlation coefficient is calculated 
directly with the raw frequency count, the calculation will be erroneous. Since log 
frequency ranges are close to the values of the familiarity ratings, we calculated the 
correlation coefficients between the log frequency and the familiarity rating. Moreover, 
from an information theory viewpoint, log-frequency can be interpreted as the 
information amount carried by a word, so our study can be interpreted as an 
investigation of the relation between the information amount and the familiarity of a 
word. In this study, we also assume the Weber-Fechner law applies to familiarity as 
noted in the Introduction.

\begin{figure}[t] 
  \begin{center}
    \includegraphics[width=8.0cm]{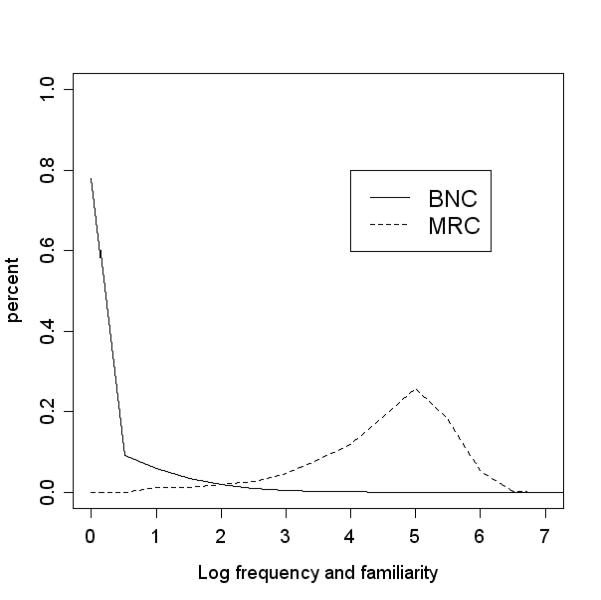}
    \caption{Distribution of MRC and BNC on a graph} 
\label{fig:distr} 
\end{center} 
\end{figure}

Before we look at the correlation coefficients, some preparatory facts need to be 
established. \figref{fig:distr} shows the distribution of MRC and the log-frequency of 
BNC. The horizontal axis shows the familiarity rating for MRC and the log-frequency 
for BNC. The vertical axis shows the distribution of words for the corresponding 
log-frequency or familiarity. The monotonically descending line shows the distribution 
of BNC, which naturally follows the LNRE tendency of natural language. In contrast, 
the MRC plot has a peak at about 5 to 6. This shows a sampling bias of the words used 
for the psycholinguistic experiment when generating the word familiarity list. In fact, 
this bias is unavoidable, since a human subject can judge familiarity only for known 
words, and so the cohesion of human judgment regarding rare words is likely to be 
lower than that for frequently used words. 

Bearing this distribution difference in mind, we analyzed the correlation of the 
familiarity lists and the log-frequency obtained from corpora.

\begin{figure}[t] \vspace*{1cm} 
  \begin{center}
    \includegraphics[width=6.5cm]{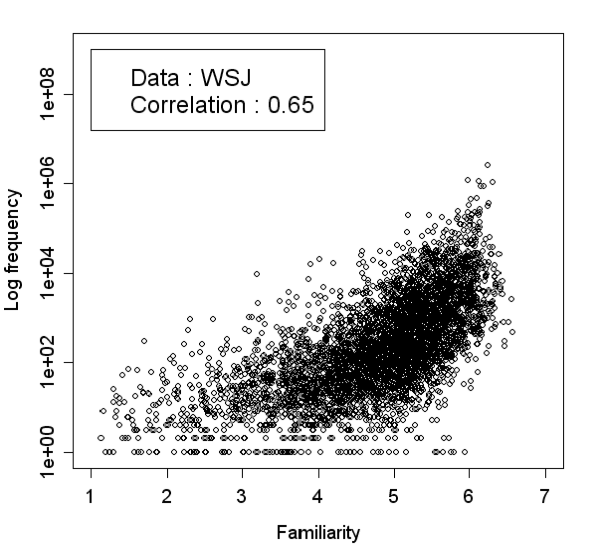}
    \includegraphics[width=6.5cm]{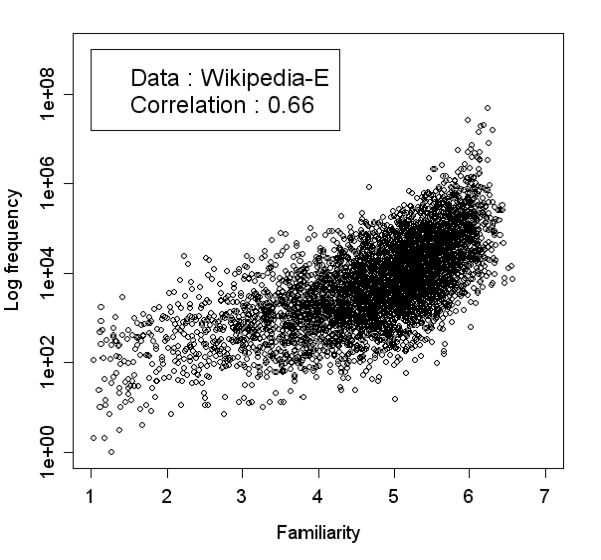}
    \includegraphics[width=6.5cm]{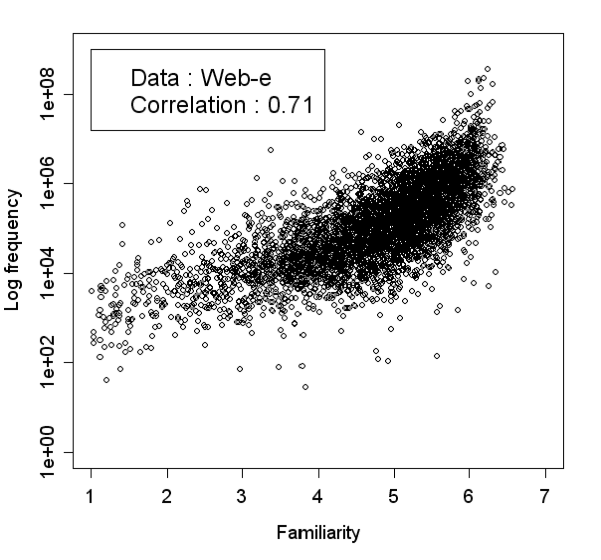}
    \includegraphics[width=6.5cm]{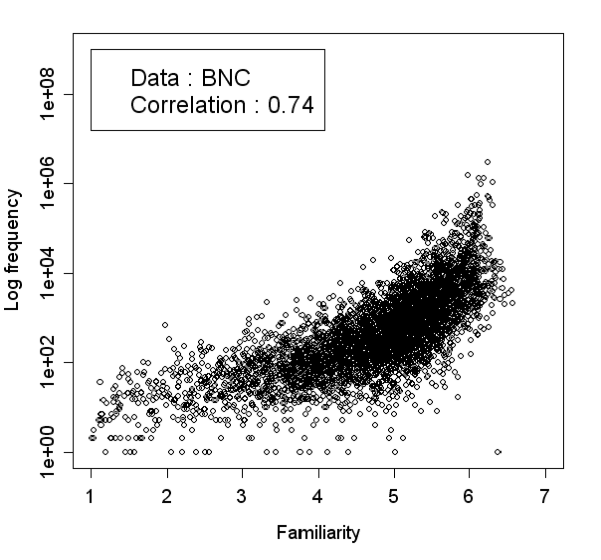}
    \includegraphics[width=6.5cm]{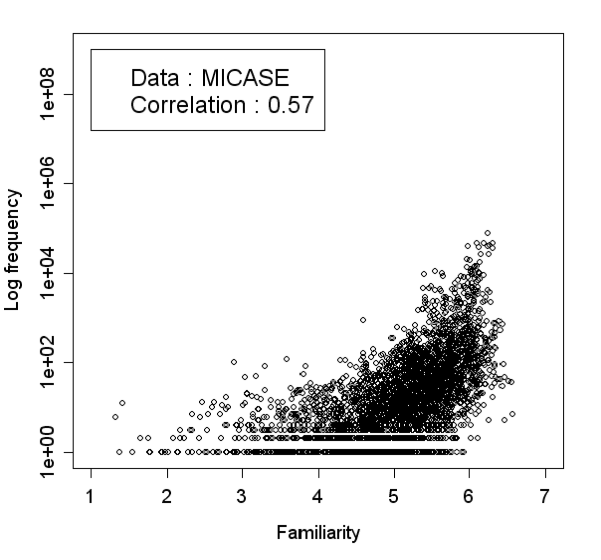}
    \caption{Word familiarity and log-frequency in English} 
\label{fig:fam-e} 
\end{center} \vspace*{2cm} 
\end{figure}

\begin{figure}[t] \vspace*{1cm} 
  \begin{center} 
    \includegraphics[width=6.5cm]{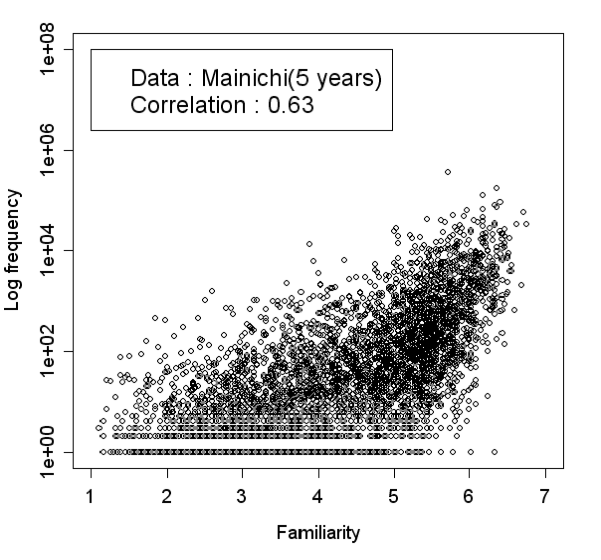} 
    \includegraphics[width=6.5cm]{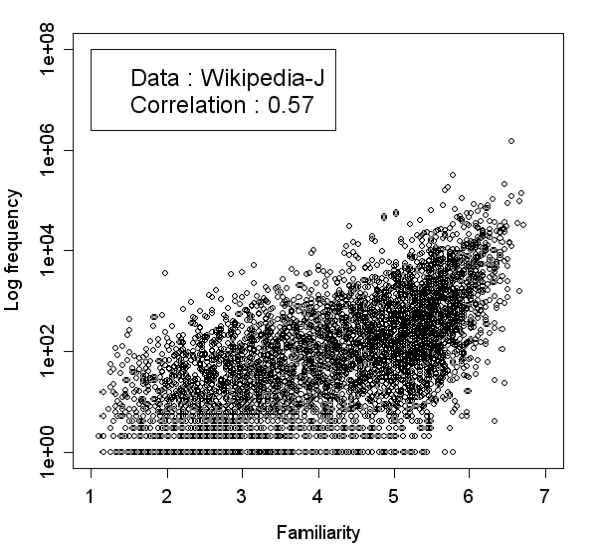}
    \includegraphics[width=6.5cm]{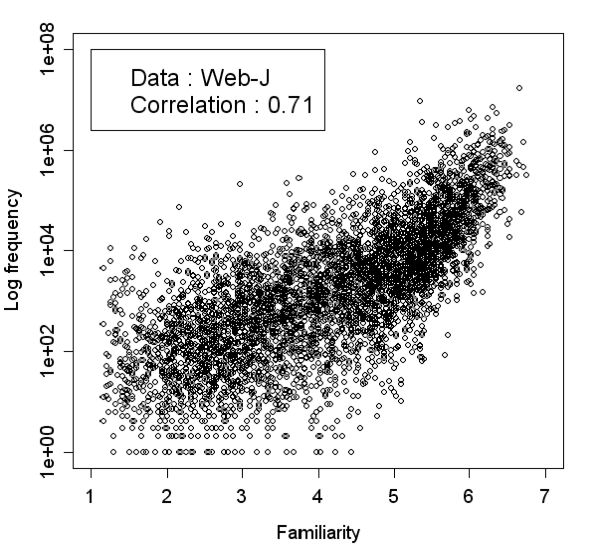}
    \includegraphics[width=6.5cm]{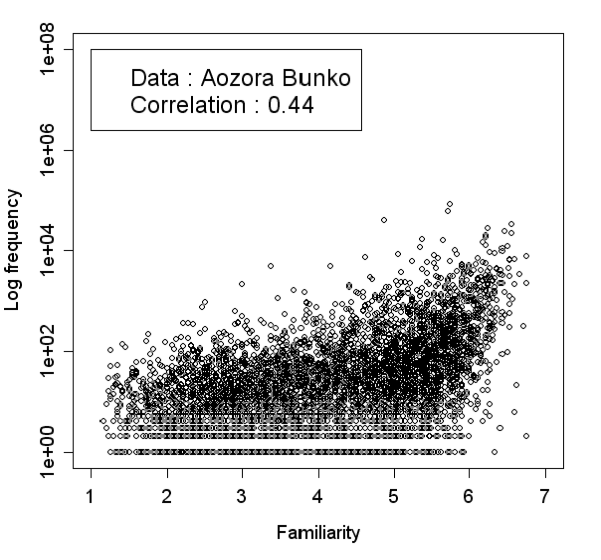}
    \includegraphics[width=6.5cm]{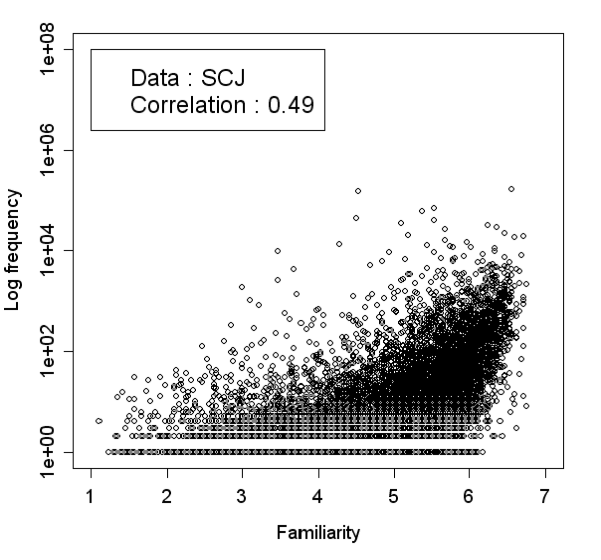}
    \caption{Word familiarity and log-frequency in Japanese} 
\label{fig:fam-j} 
\end{center} \vspace*{2cm} 
\end{figure}

The correlation plots of word familiarity and frequency are shown in \figref{fig:fam-e} 
for English and in \figref{fig:fam-j} for Japanese. Each graph corresponds to a corpus, 
so there are ten graphs in total, five for each language. The horizontal axes show the 
familiarity rating, whereas the vertical axes show the log-frequency. Each plot 
corresponds to a word for which a familiarity rating and a log-frequency were acquired.

The plots in general move from the bottom-left to the upper-right, showing the global 
trend of correlation. Moreover, the plots form a fat cloud or take a triangular form with 
the largest angle in the lower right corner. Such a triangular form indicates the inclusion 
of more rare but familiar words. The triangular tendency is most prominent when the 
data size is small and the familiarity list is large. For example, the graphs for Japanese, 
except for Web-J, show this triangular tendency, but the tendency resolves into a cloud 
without the angle when the corpus size is increased. Taking BNC as an example, five 
rare and familiar words are listed as follows: 
\begin{quote} 
spank (familiarity rating=5.36 / frequency=19), pimple (5.57 / 20), dime (5.86 / 39), 
easygoing (5.25 / 41), quart (5.68 / 50) 
\end{quote} 
As shown, these words are familiar, but are unlikely to appear frequently in a corpus 
depending on the content. 

On the other hand, the fact that there are very few plots in the upper left area shows 
there are few unfamiliar but frequent words. The tendency for plots to exist only in the 
lower right half of the graph indicates that, for a given corpus, frequent words are 
always familiar, whereas familiar words are not necessarily frequent. More precisely, for 
a given corpus, high frequency {\em is a necessary condition} for a word to attain a 
high familiarity rating, but {\em is not a sufficient condition} to make the word familiar. 
A question is then raised: what is a sufficient condition to make a familiar word 
frequent? Since the investigation was done for a given corpus, the answer to this 
question depends on the corpus, and thus a more precise question is: what is the corpus 
condition under which a familiar word is used frequently?

\begin{table}[t] 
\begin{center} 
\caption{Coverage and Correlation Coefficients} 
\label{tab:correlation} 
\begin{tabular}{|l|p{3cm}|p{3cm}|p{3cm}|} 
\hline 
label &Coverage of the words in the familiarity list & Pearson's correlation coefficient & 
Spearman's correlation coefficient \\
\hline 
WSJ (3-years) & 4772 (97.507\%) & 0.6520 & 0.7082 \\
Wikipedia-E & 4882 (99.755\%) & 0.6677 & 0.6801 \\
Web-E & 4892 (99.959\%) & 0.7185 & 0.7359 \\
BNC & 4869 (99.489\%) & 0.7438 & 0.7776 \\
MICASE & 3608 (73.722\%) & 0.5744 & 0.7127 \\
\hline 
\hline 
Mainichi (5 years) & 38337 (55.926\%) & 0.6327 & 0.5330 \\
Wikipedia-J & 44784 (65.330\%) & 0.5737 & 0.4452 \\
Web-J & 48469 (70.706\%) & 0.7193 & 0.4920 \\
Aozora & 43430 (63.335\%) & 0.4484 & 0.3451 \\
SCJ & 19736 (28.625\%) & 0.4931 & 0.5097 \\
\hline 
\end{tabular} 
\end{center} 
\end{table}

\begin{figure}[t] 
  \begin{center}
    \includegraphics[width=7cm]{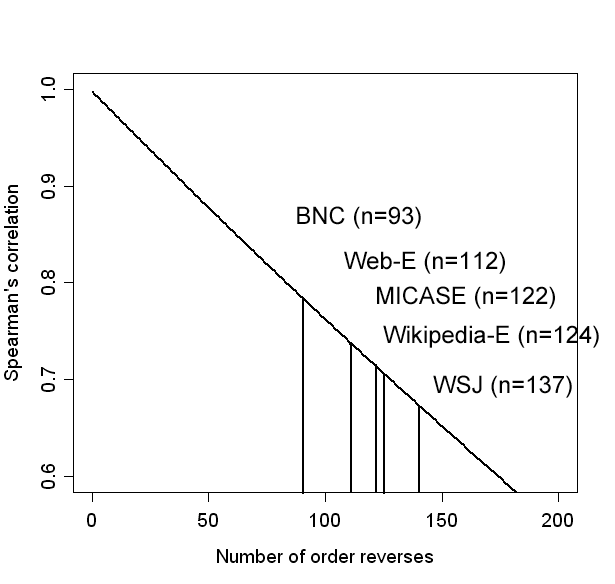} 
      \includegraphics[width=7cm]{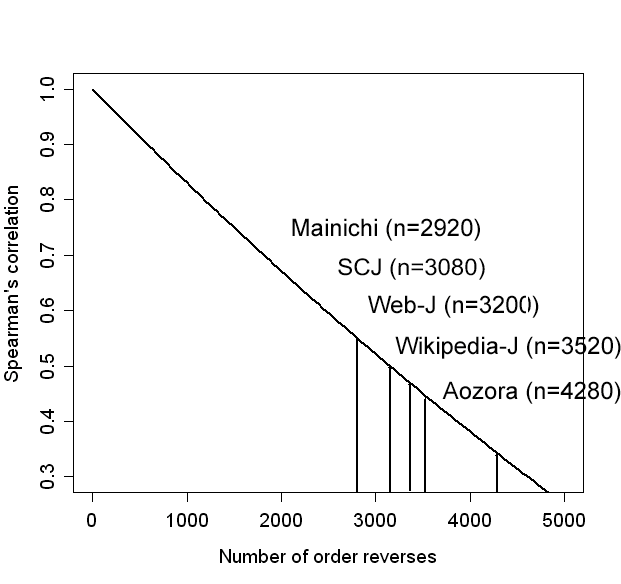} 
\caption{Number of ranking reverses and Spearman's correlation: in English (left) and 
Japanese (right)} 
\label{fig:simulation} 
\end{center} 
\end{figure}

To verify the correlation degree, we calculated the correlation coefficients as shown in 
\tabref{tab:correlation}. The table shows three values for each corpus: 
\begin{description} 
\item[Coverage:] the number (and percentage) of the words in the familiarity list found 
in the corpus, 
\item[Pearson:] Pearson's product-moment correlation coefficient, and 
\item[Spearman:] Spearman's rank-order correlation coefficient. 
\end{description} 
Note that {\em missing values} ---words which are in the familiarity list but {\em not} 
in the corpus--- were not considered when calculating Pearson's coefficient, whereas 
they were taken into consideration when calculating Spearman's coefficient.

The coverage (first column) shows that the larger the corpus size, the greater the 
coverage. Since the magnitude of MRC is in the thousands, it was well covered: only 
two words from MRC were missing from the Web-E corpus. In contrast, since the 
Japanese familiarity list is large, even when using Web-J, the coverage only amounted 
to about 70\%; therefore, the number of missing values is large in the case of Japanese.

The second and third columns show the two correlation coefficients. The general 
tendency is that a corpus with a high Pearson coefficient has a high Spearman 
coefficient. For example, the highest correlation coefficients of both types were attained 
with BNC for English. Still, there are inconsistencies; for example, Web-J had the 
highest Pearson coefficient for Japanese, whereas Mainichi had the highest Spearman 
coefficient.

How significant these differences in the correlation coefficients are can be intuitively 
understood through simulation as follows. For each English and Japanese familiarity list, 
a copy of a list is generated (which of course perfectly correlates with the original). For 
this copy, two successively ranked words are randomly selected, and then their orders 
are reversed and the correlation with the original list is measured. By repeating this 
reversing procedure, the correlation will decrease as in \figref{fig:simulation}. The 
horizontal axes show the number of reversed pairs, and vertical axes show the 
correlation. In the English case, starting from 4894 ranked words, it required only 93 
reverses to attain the BNC correlation, and 124 reverses to reach the Wikipedia-E level. 
In Japanese, since the Amano list is about ten times larger, more reverses are required to 
decrease the correlation from Mainichi to Aozora. For the 68,550 words of the Amano 
list, only 1340 additional reverses are needed to decrease the Mainichi correlation level 
to that of Aozora. Correlation thus decreases quickly even with slight inconsistencies 
with the original. Therefore, although the correlation seems different from 
\tabref{tab:correlation}, the difference is not as significant as it might seem.

To further reveal differences among the corpora, \tabref{tab:data-correlation} shows 
the correlation among the corpora for words in the familiarity list. As shown in the first 
block of \tabref{tab:data-correlation}, many of the correlation values are over 0.8, 
some even near 1.0, showing high correlation in English. For Japanese, where the 
Amano list has 68,550 words, the values are naturally lower than in the English case, 
but still consistently above 0.34: this corresponds to almost 4280 reverses (from the 
Aozora correlation being almost 0.34, which corresponds to the 4280 reverses in 
\figref{fig:simulation}). Consequently, the difference is not as significant as it might 
seem from the coefficient values.

\begin{table*}[t] 
\begin{center} 
\caption{Pearson/Spearman correlation coefficients among corpora: English (first 
block) and Japanese (second block)} 
\label{tab:data-correlation} 
\begin{tabular}{|l|l|l|l|l|l|} 
\hline & \multicolumn{5}{|c|}{Correlation with MRC List for English} \\
\hline & WSJ & Wikipedia-E & Web-E & BNC & MICASE \\
\hline WSJ &\ 1.0 / 1.0 & 0.96 / 0.87 & 0.90 / 0.83 & 0.97 / 0.89 & 0.81 / 0.81 \\
Wikipedia-E & --- &\ 1.0 / 1.0 & 0.88 / 0.87 & 0.97 / 0.90 & 0.77 / 0.82 \\
Web-E & --- & --- &\ 1.0 / 1.0 & 0.92 / 0.86 & 0.81 / 0.79 \\
BNC & --- & --- & --- &\ 1.0 / 1.0 & 0.87 / 0.85 \\
MICASE & --- & --- & --- & --- & \ 1.0 / 1.0 \\
\hline 
\hline & \multicolumn{5}{|c|}{Correlation Among Whole Amano List} \\
\hline & Mainichi & Wikipedia-J & Web-J & Aozora & SCJ \\
\hline Mainichi &\ 1.0 / 1.0 & 0.52 / 0.63 & 0.75 / 0.63 & 0.34 / 0.46 & 0.35 / 0.55 \\
Wikipedia-J & --- &\ 1.0 / 1.0 & 0.52 / 0.60 & 0.73 / 0.45 & 0.58 / 0.52 \\
Web-J & --- & --- &\ 1.0 / 1.0 & 0.47 / 0.78 & 0.35 / 0.56 \\
Aozora & --- & --- & --- &\ 1.0 / 1.0 & 0.73 / 0.45 \\
SCJ & --- & --- & --- & --- &\ 1.0 / 1.0 \\
\hline 
\end{tabular} 
\end{center} 
\end{table*}

Still, the differences among corpora remain a question: what conditions cause one 
corpus to be more closely correlated with familiarity than another corpora. Therefore, in 
the remainder of this article we consider what these conditions could be and verify that 
two corpus criteria affect correlation with the familiarity list.

The correlation scores show that the correlation rises as the amount of data increases. 
This is natural, since, for a given corpus, high frequency is a necessary condition for 
high familiarity, and a larger corpus would increase the sampling size. Indeed, 
\tabref{tab:correlation} shows that the correlation of web data was among the highest if 
measured by the Pearson coefficient. Moreover, the larger the amount of data, the less 
triangular the plots are in \figref{fig:fam-e} and \figref{fig:fam-j}, which reflects 
increased correlation. Therefore, the data size seems to account for some of the 
correlation increase.

The difference is caused not only because of the size counts, though. For example, 
\tabref{tab:correlation} shows that for English, BNC beats Web-E in terms of both 
correlation coefficients, yet BNC is smaller in magnitude than Web-E. Similarly, for 
Japanese, Mainichi beats Wikipedia-J for both coefficients, despite Mainichi being 
much smaller than Wikipedia-J.

Such differences are probably due to differences in the corpus domains. Especially, 
when we compare the corpora with higher correlation (BNC, web, newspapers) and 
those with lower correlation (Aozora, Wikipedia), we see that the corpora containing 
more daily content seem to have higher correlation. BNC includes a spoken part (10\%) 
which consists of orthographic transcriptions of unscripted informal conversations 
(recorded by volunteers selected from different age, regional and social classes in a 
demographically balanced way) and spoken language collected in different contexts, 
ranging from formal business or government meetings to radio shows and phone-ins. 
Also, MICASE and SCJ ---both spoken--- have relatively high Spearman coefficients 
despite their size. In contrast, Aozora and Wikipedia are at the extreme of written form. 
The highly correlated corpora include texts closer to spoken language, whereas the less 
correlated ones tend to be formed only of written language.

Consequently, the answer to the question raised in this section appears to consist of two 
factors: 
\begin{itemize} 
\item 
corpus size: a larger corpus better correlates with familiarity ratings 
\item 
corpus domain: spoken correlates better than written 
\end{itemize} 
In the following two sections, we verify the significance of each of these factors. 

\begin{figure}[t] 
\begin{minipage}{0.45\textwidth} 
  \begin{center}
    \includegraphics[width=7.5cm]{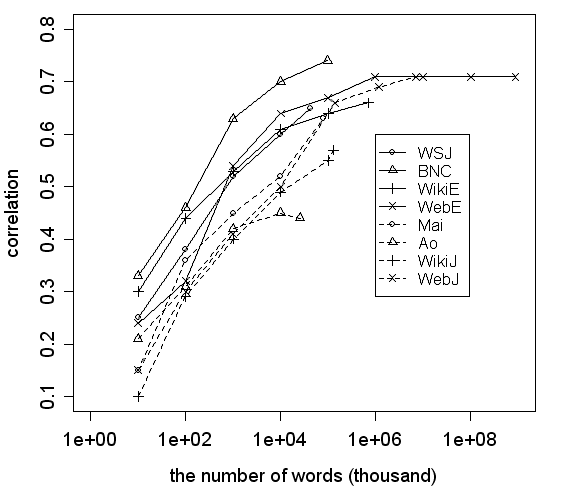} 
\caption{Data size (number of words, in thousands) and correlation} 
\label{fig:growth} 
\end{center} 
\end{minipage} \hspace*{1cm} 
\begin{minipage}{0.45\textwidth} 
  \begin{center}
    \includegraphics[width=7.5cm]{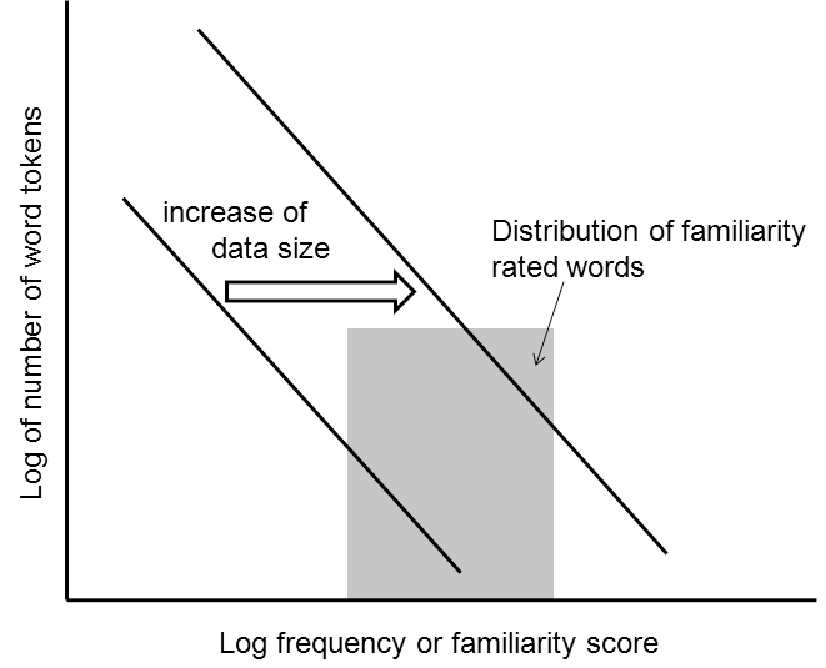} 
\caption{Corpus size and correlation} 
\label{fig:model} 
\end{center} 
\end{minipage} 
\end{figure}

\section{Effect of Data Size} 
\label{sec:size}
Each corpus was divided into parts with the number of words in each part increasing 
exponentially (i.e., the number of words was 10${}^1$, 10${}^2$, 10${}^3 \ldots$) 
until the maximum data size was reached, where a larger data set included the smaller 
ones. The relation between data size and correlation is shown in \figref{fig:growth}. 
The horizontal axis shows the number of words (in thousands) in log-scale, and the 
vertical axis shows the correlation. Nine lines are shown, each corresponding to a 
corpus (except for MICASE, which was available online only). English corpora are 
represented by solid lines and Japanese corpora by dashed lines. Since the data size 
differs according to the corpus, the line lengths differ.

For every corpus, the correlation increased with size. The increase shows a log-linear 
tendency up to about 1 billion words, where a plateau is reached. The effect of data size 
on correlation is clear from this data.

Such an increase in correlation with data size can be intuitively explained as follows. As 
shown in \figref{fig:model}, consider a graph where the horizontal axis is the 
log-frequency and the vertical axis is the logarithm of the number of words (in 
\figref{fig:distr}, the vertical axis was the distribution, but here it is a histogram). 
Every corpus follows the power-law with its gradient being almost the same for each 
natural language text. For a given type of data, an increased amount of data causes a 
shift of the line towards the upper right. Since the distribution of word familiarity is 
biased towards the highly frequent range (indicated by the gray zone in the figure), the 
larger the data, the better sampled the words in the familiarity list are. Thus, for any 
given data type, increased data size raises the correlation.

\section{Effect of Domain} 
\label{sec:domain}
Having seen that a larger amount of data strengthens the correlation, the remaining 
question is what causes a corpus of a given size to correlate more strongly with the 
familiarity ratings. This corresponds to finding the reason for the vertical margin among 
the lines in \figref{fig:growth} when the data size was fixed.

In our quest for a statistical explanation for this margin, we examined several statistics 
for each corpus of the same size. The statistics included the percentage of covered 
words in the corpus, the percentage of covered tokens in the corpus, the entropy among 
covered words, the frequency distribution of covered words, and the distribution of 
bigram and trigram numbers for covered words. None of these statistics were consistent 
with the degree of correlation. The differences between the corpora lie in the different 
appearance of individual words depending on the domain.

Thus, for each corpus, we examined words which contributed to low correlation. For a 
given corpus, we considered common words found in the corpus and the familiarity list. 
The corpus gives a ranking $n$ and the familiarity list gives a ranking $m$ to every 
word. Words with large absolute values of $n-m$ are the words which contribute most 
to lowering the correlation coefficient. Words with the highest $m-n$ are shown in the 
first block of \tabref{tab:diff}, whereas words of the highest $n-m$ are shown in the 
second block. The first block includes more daily usage and spoken words, whereas the 
second block includes words more often found in written text. For example, clothing 
and diet words in the first block include words typically used in daily communication, 
whereas the English words borrowed from French in the second block are more likely to 
be written. In Japanese, the tendency of typically written words to be among the large 
$n-m$ words is even more clear, since the Amano list is larger.

\begin{table} 
\begin{center} 
\caption{Words with the largest ranking difference in the corpus and the familiarity 
list} 
\label{tab:diff} 
\begin{tabular}{|l|l|l|l|l|} 
\hline \multicolumn{5}{|c|}{corpus: low, MRC: high } \\
\hline WSJ & Wikipedia-E & Web-E & BNC & MICASE \\
\hline pencil & mileage & doughnut & sock & bedroom \\  
noisy & towel & coke & nickel & thirsty \\
oven & thoughtful & steady & boring & spoon \\
happiness & bra & hunger & shrimp & sunshine \\
\hline 
\hline \multicolumn{5}{|c|}{corpus: high, MRC: low } \\
\hline lire & sonata & dell & essence & hypothesis \\
southland & hank & portal & lorry & velocity \\
gore & aurora & fort & debut & precipitate \\
charter & belle & enterprise & rover & mass \\
\hline 
\end{tabular} 
\end{center} 
\end{table}

\begin{figure}[t] 
\begin{minipage}{0.45\textwidth} 
  \begin{center}
    \includegraphics[width=7.5cm]{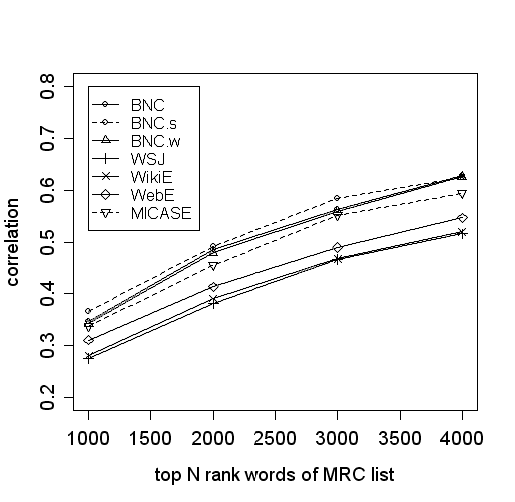} 
\caption{Top $N$ Rank Words of MRC and Correlation} 
\label{fig:rankmrc} 
\end{center} 
\end{minipage} \hspace*{1cm} 
\begin{minipage}{0.45\textwidth} 
  \begin{center}
    \includegraphics[width=7.5cm]{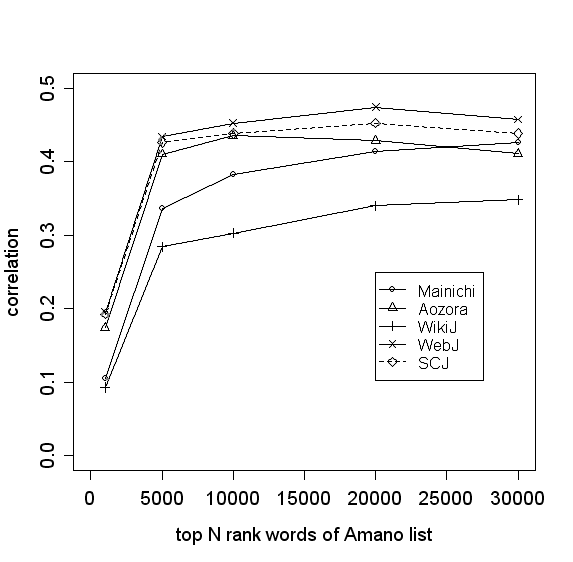} 
\caption{Top $N$ Rank Words of Amano list and Correlation} 
\label{fig:rankmrc-j} 
\end{center} 
\end{minipage} 
\end{figure}

We next compared the correlation coefficients with spoken and written corpora. Since 
the corpus size will affect the correlation, the same quantity of $K$ words was taken 
from each corpus, and the correlation of this sample with the familiarity list was 
calculated. Since the corpus size differed, $K$ was defined as the smallest number of 
words among the corpora in each language (thus, MICASE for English where 
$K_{eng}$=1,279,792, and SCJ for Japanese where $K_{jap}$=7,498,763 words). 
Since the number of words in the familiarity list also affects the correlation, the top 
$N$ ranked words were taken from the familiarity list, and the correlation with each 
corpus of size $K$ was calculated (thus, $K$ is a constant, whereas $N$ is a parameter 
for each language). \figref{fig:rankmrc} shows the results for English and 
\figref{fig:rankmrc-j} shows those for Japanese. The horizontal axis shows $N$, while 
the vertical axis shows the correlation.

For English, since we were interested in spoken versus written corpora, BNC was 
separated into a BNC-spoken part (denoted BNC-s in the figure) and a BNC-written 
part (BNC-w)\footnote{BNC was originally tagged to show whether the text is spoken 
or written.}. The correlation was calculated also for BNC as a mixture of spoken and 
written with size $K_{eng}$. Thus, there are seven lines in total, two corpora of spoken 
text (dashed lines) and five corpora of written (solid lines). For Japanese, there are five 
lines, only one of which is dashed. Note that the ranges of the horizontal axes differ due 
to the familiarity list size: for MRC the range goes to 4000 and for the Amano list it 
goes to 30,000.

In general, the dashed lines are above most of the solid lines. This shows the tendency 
of the spoken corpus to correlate better with the familiarity list than the written corpus 
does. Still, some solid lines were above the dashed lines in the two graphs. For English, 
the BNC lines were higher than that of MICASE. The reason for this could not be 
identified and remains to be determined in our future work, but a possible reason lies in 
the balanced nature of BNC. BNC is characterized in two ways regarding familiarity: it 
contains spoken data, and it was constructed to represent the English of daily usage. 
With regard to the first characteristic of BNC, the MICASE content is academic speech, 
so even though MICASE consists of spoken language it contains much less familiar 
content. In contrast, BNC-spoken is more general. Consequently, if a large-scale corpus 
of daily spoken language is constructed, it should correlate more strongly with MRC. 

Regarding the second characteristic of BNC, that it is a representative collection of 
contemporary language, the fact that BNC-written is more highly correlated than 
MICASE shows that the content of BNC closely matches the words in the MRC list. 
Even to the present day, BNC is the only corpus developed with an explicit strategy to 
create a representative collection of printed English that is available and large in scale. 
A good contrast is Web-E, which also includes various publications of the day, but is 
matched less well with MRC. This shows that the familiarity rating is formed of more 
controlled, meaningful data rather than with noisy, uncontrolled data.

Verification through this approach in Japanese is limited, though, since there is no 
BNC-like corpus in any language other than English. In Japanese, SCJ, which consists 
of university lectures, had lower correlation than Web-J. Thus, Web-J might be a fairly 
good replacement for BNC. Web-E's correlation line was among those of the corpora 
based on written texts, so there seems to be a qualitative difference between Web-E and 
Web-J. This could be due to the fact that Japanese web pages are written and read 
mostly by native speakers of Japanese, whereas English web pages are often written and 
read by non-native speakers. Thus, if English web-pages could be filtered to obtain only 
those written by native speakers of English, a higher correlation with MRC might be 
attained. However, since web pages are noisy sources of data, the construction of a 
Japanese corpus that is ``representative of the language'' still lies in the future.

Comparing \figref{fig:rankmrc} and \figref{fig:rankmrc-j} leads to another 
observation. For Japanese, the correlation levels off when the top familiarity list reaches 
5000 words. This suggests that the familiarity rating is only meaningful with respect to 
the corpus frequency up to about 5000 words. Thus, an MRC list of an order of about 
5000 words seems adequate, if there is no language difference. The Dale-Chall list 
\citep{chall95} used in reading level assessment consists of 3000 words, which seems 
reasonable according to the results of this experiment.

Is frequency correlated with familiarity? Our correlation analysis on corpora suggests 
that log frequency ---as the information amount carried by a word under 
Weber-Fechner's law--- does correlate with the familiarity rating, if measured on a 
gigantic corpus of spoken representative language of the day. This is supported by the 
following observations made so far: 
\begin{itemize} 
\item 
Log-frequency measured in written corpora correlates fairly well with the familiarity 
ratings. 
\item 
The correlation is higher for spoken language corpora than for written corpora. 
\item 
A corpus of academic speech was less correlated than the BNC-written corpus, which 
comprises representative language of the day. 
\end{itemize} 
Unfortunately, such corpora do not exist in any language, although the closest is BNC. 
Therefore, a final conclusion must wait until such a corpus becomes available. Further 
investigation to verify the relationship between frequency and familiarity remains as our 
future work.

\section{Conclusion}
Even though it has been assumed that word frequency correlates with word familiarity, 
how strong this correlation is has not been thoroughly investigated. In this paper, we 
report on our analysis of various corpora in English and Japanese.

The correlation coefficient of corpus log-frequency with a word familiarity list was 0.57 
to 0.74 for English using the MRC familiarity list, and 0.45 to 0.72 for Japanese with 
the Amano list. For a given corpus, frequent words always had high familiarity, but 
familiar words did not necessarily have high frequency.

To explain why the log-frequency of some corpora was better correlated with the 
familiarity rating than was the case for other corpora, we investigated two conditions. 
The first was the corpus size. The log-frequency of larger corpora was more strongly 
correlated with familiarity ratings than that of smaller corpora. By changing the corpus 
size, we found that increasing the amount of data will increase the correlation in a 
log-linear manner up to 1 billion corpus words. The second condition was the type of 
corpus. The log-frequency was more strongly correlated with the familiarity ratings 
when the corpus consisted of spoken rather than written data. Even when the corpus 
content was academic speech, the log-frequency of the corpus was more highly 
correlated with the familiarity ratings than that of most written corpora. Such findings 
partly show the nature of familiarity as the information amount carried by the word, 
under a more general model of Weber-Fechner's law that the relationship between 
stimulus and perception is in general logarithmic.

From the language engineering viewpoint, a log-frequency list ---if obtained from a 
large-scale corpus--- could be used for a pseudo-measurement of familiarity scores. The 
approximation will be better if collected from spoken language and daily content. Since 
most current corpora are collections of written text, our work also suggests that the 
construction of corpora consisting of typical spoken text (e.g., not university lectures) 
will be useful, as will the collection of texts for BNC-type corpora in various languages. 
How such pseudo-scores can be applied to reading level assessment will be part of our 
future work.

\bibliographystyle{natbib} 
\bibliography{fam} 
\end{document}